\documentclass[conference]{IEEEtran}
\IEEEoverridecommandlockouts
\usepackage{cite}
\usepackage{amsmath,amssymb,amsfonts}
\usepackage{algorithmic}
\usepackage{graphicx}
\usepackage{textcomp}
\usepackage{xcolor}
\usepackage{multirow}
\usepackage{floatrow}
\def\BibTeX{{\rm B\kern-.05em{\sc i\kern-.025em b}\kern-.08em
    T\kern-.1667em\lower.7ex\hbox{E}\kern-.125emX}}
\usepackage{physics}
\usepackage{amsmath}
\usepackage{tikz}
\usepackage{mathdots}
\usepackage{yhmath}
\usepackage{cancel}
\usepackage{color}
\usepackage{siunitx}
\usepackage{array}
\usepackage{multirow}
\usepackage{amssymb}
\usepackage{gensymb}
\usepackage{tabularx}
\usepackage{extarrows}
\usepackage{booktabs}
\usetikzlibrary{fadings}
\usetikzlibrary{patterns}
\usetikzlibrary{shadows.blur}
\usetikzlibrary{shapes}
\makeatletter
\newcommand{\linebreakand}{%
  \end{@IEEEauthorhalign}
  \hfill\mbox{}\par
  \mbox{}\hfill\begin{@IEEEauthorhalign}
}
\makeatother

\begin{document}

\title{ An Asymmetric Loss with Anomaly Detection LSTM Framework for Power Consumption Prediction\\
}

\author{\IEEEauthorblockN{Jihan Ghanim}
\IEEEauthorblockA{\textit{Dept. of Electrical and Computer Engineering} \\
\textit{American University of Beirut}\\
Beirut, Lebanon \\
jhg05@mail.aub.edu}
\and
\IEEEauthorblockN{Maha Issa}
\IEEEauthorblockA{\textit{Dept. of Electrical and Computer Engineering} \\
\textit{American University of Beirut}\\
Beirut, Lebanon \\
mgi03@mail.aub.edu}
\linebreakand 
\IEEEauthorblockN{Mariette Awad}
\IEEEauthorblockA{\textit{Dept. of Electrical and Computer Engineering} \\
\textit{American University of Beirut}\\
Beirut, Lebanon \\
mariette.awad@aub.edu.lb}
%\and
%\and
%\and
}

\maketitle

\begin{abstract}

%An accurate load forecasting model,  having minimal under predictions, can prevent any undesired power outages due to under productions of electricity. It can also ensure a balance between the energy supply and the energy demand without unnecessary over productions.
Building an accurate load forecasting model with minimal underpredictions is vital to prevent any undesired power outages due to underproduction of electricity. However, the power consumption patterns of the residential sector contain fluctuations and anomalies making them challenging to predict. In this paper, we propose multiple Long Short-Term Memory (LSTM) frameworks with different asymmetric loss functions to impose a higher penalty on underpredictions. We also apply a density-based spatial clustering of applications with noise (DBSCAN) anomaly detection approach, prior to the load forecasting task, to remove any present oultiers. %Also, to consider the seasonality trend, seasonality splitting is performed on the two considered German and French residential multi-variant datasets containing hourly power consumption, weather, and calendar features.
Considering the effect of weather and social factors, seasonality splitting is performed on the three considered datasets from France, Germany, and Hungary containing hourly power consumption, weather, and calendar features.
%First, two  German and French residential multi-variant datasets, having hourly power consumption, weather, and calendar features, are split into three seasonal datasets each to consider the seasonality trend. After that, anomaly detection and removal are performed  on the hourly power consumptions of the seasonal data. Then, these seasonal datasets are separately fed to the proposed LSTM models. At last, the LSTM models are evaluated using the root-mean-square error (RMSE) metric.
Root-mean-square error (RMSE) results show that removing the anomalies efficiently reduces the underestimation and overestimation errors in all the seasonal datasets. Additionally, asymmetric loss functions and seasonality splitting effectively minimize underestimations despite increasing the overestimation error to some degree.
%Finally, seasonality splitting proves to help in decreasing  underpredictions while slightly increasing overpredictions. 
Reducing underpredictions of electricity consumption is essential to prevent power outages that can be damaging to the community.
\end{abstract}

\begin{IEEEkeywords}
Anomaly detection, asymmetric loss, DBSCAN, LSTM, seasonality.
\end{IEEEkeywords}

\section{Introduction}
\label{Introduction}
%Over the past decade, electricity consumption has drastically increased. The reduction in natural resources, such as fuel, and the environmental pollution caused by conventional power plants have resulted in massive attempts to prevent overproduction of electricity \cite{article1}. 
Over the past decade, electricity consumption has drastically increased, leading to a reduction in natural resources and an aggravation in the environmental pollution. Accordingly, massive attempts have been undertaken to prevent overproduction of electricity \cite{article1}. Despite the importance of preventing overproduction, avoiding underproduction of electricity remains crucial since it may lead to power outages and electricity blackouts, which can decline the economic and industrial development \cite{article2} and even result in terrible situations. For instance, in 2021, an intense power outage occurred in Lebanon, putting hospitals and essential services in real crisis. To avoid underproduction and overproduction of electricity, power companies are continuously investigating a better load forecast and a more reliable plan of energy usage \cite{article3}.

Prior work in power consumption estimation used various machine learning techniques such as Recurrent Neural Networks (RNN) \cite{article8}, Long Short-Term Memory (LSTM) algorithm \cite{article3,article6,article7}, or a hybrid approach using both Artificial Neural Network (ANN) and Support Vector Machine (SVM) methods \cite{article11}. Generally, the cost functions that machine learning models aim to minimize are, by default, symmetric. Such loss functions typically perform well in applications where underestimations and overestimations have similar effects. However, in some applications, underestimations may have more unpleasant consequences in comparison to overestimations, which encouraged many researchers to build models with asymmetric objective functions for a better flexibility \cite{article22,article23,article24,article25,article26,6403624}. This asymmetry assigns more weight to the predictions that researchers are more interested in minimizing, which are the underpredictions in the power consumption estimation case. The ideal scenario would be to keep the generated power slightly higher than the power demand to prevent an undesired power outage or an emergency purchase of power \cite{article26, 6403624}.

Power consumption forecasting is challenging due to the presence of fluctuations in the power consumption patterns, especially in the residential sector \cite{article6}. Such fluctuations are affected by weather and social conditions \cite{article12}. 
%Nevertheless, data gathering errors, wrong measurements, and sometimes special events also lead to extreme values that are sources of unusual patterns or anomalies \cite{10.5555/3001460.3001507}.
Power consumption patterns may also contain unusual values or anomalies due to data gathering errors, wrong measurements, and sometimes special events \cite{10.5555/3001460.3001507}.
Researchers in \cite{article8} stated that the presence of anomalies has increased the model’s error.

\begin{figure*}[htbp]
    \centering
\includegraphics[scale=0.61]{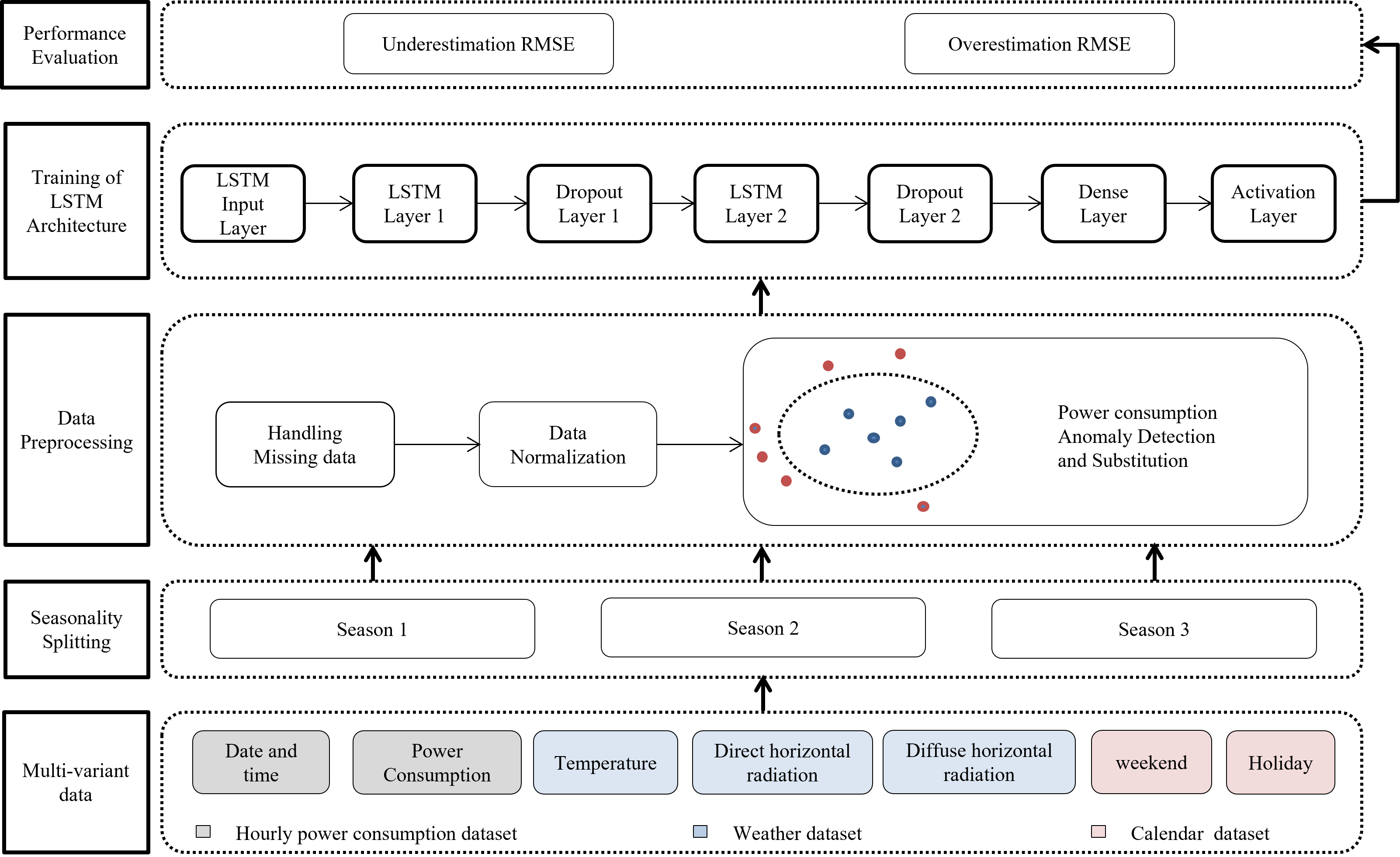}
    \caption{System framework for power consumption estimation.}
    \label{system_framewok}
\end{figure*}

In this work, we aim to improve the power consumption forecasting in the residential sector by eliminating outliers and reducing errors in predictions, mainly in underpredictions. An illustration of the proposed framework is shown in ``Fig.~\ref{system_framewok}''. First, we combine the hourly power consumption datasets with their corresponding weather and calendar datasets\cite{article16}. To consider the seasonality factor, we then split each multi-feature dataset into three seasonal datasets. Then, anomaly detection is applied on the power consumption feature using density-based spatial clustering of applications with noise (DBSCAN) algorithm \cite{10.5555/3001460.3001507}. The detected outliers are substituted with more realistic values to obtain better results. Finally, the seasonal datasets are separately fed to three various LSTM models: a regular LSTM model with a symmetric loss function, and two LSTM models each with a unique asymmetric loss function. 
% to study the effect of outliers on the load predictions. 
%Two LSTM models with two different asymmetric loss functions are implemented and compared to a LSTM model with a symmetric loss function. 
These models are evaluated based on both the underestimation root-mean-square error (RMSE) and the overestimation RMSE. % to examine the asymmetry impact on producing more accurate forecasts with limited underpredictions. %An illustration of this process is shown in ``Fig.~\ref{system_framewok}''.
Results show that the proposed anomaly detection and substitution technique minimized both the overprediction and underprediction errors in all the seasonal datasets. Additionally, the proposed asymmetric loss functions and the seasonality splitting proved their success in limiting the underestimation error, but they increased the overestimation error in comparison to the symmetric loss. The contributions of this work are: (1) examining the impact of various asymmetric loss functions embedded in a LSTM architecture on limiting underpredictions and producing better load forecasts in the presence and absence of outliers and (2) investigating the impact of seasonality on the load estimation task.

The remainder of this paper is organized as follows: Section~\ref{Related Work} presents the related work, Section~\ref{Methodology} formulates the methodology, and Section~\ref{results} describes the conducted experiments and reports the obtained results. Finally, conclusions, limitations, and future work are presented in Section~\ref{Conclusion}.

%\begin{figure*}[htbp]
%    \centering
%\includegraphics[scale=0.13]{diagram.png}
%    \caption{System framework for power consumption estimation.}
 %   \label{system_framewok}
%\end{figure*}

\section{Related Work}
\label{Related Work}
The concept of asymmetric loss was discussed by several researchers especially in forecasting applications \cite{article22,article23,article24,article25,article26,6403624}. An asymmetric linear-linear cost for a load forecasting framework was introduced by \cite{article23} based on support vector regression (SVR). The applied insensitive cost consists of different penalties for overpredictions and underpredictions, which reduced the economic cost by a range of 42.19\% to 57.39\% \cite{article23}. In \cite{article26}, a quadratic and asymmetric loss function based on SVR was used to reduce the underestimates in load forecasting. The underpredictions decreased from 50\% to an average of 1.91\% with a minimal increase of 0.3\% on average in the error rate \cite{article26}. Another cost-oriented approach for load forecasting applied a differentiable piece-wise loss based on Huber function \cite{article25}. This cost-oriented framework was integrated on ANN and Multiple Linear Regression (MLR) showing promising results in reflecting real costs especially in ANN with an improvement of 13.74\% compared to regular mean squared error (MSE) \cite{article25}. In \cite{article22}, an asymmetric loss function was proposed to reflect different costs of fault detection and premature maintenance in predictive maintenance scenarios. The loss, consisting of a linear and an exponential function for overestimates and underestimates respectively, produced a more representative machine learning model of real business cases \cite{article22}. Authors in \cite{article24} evaluated different types of asymmetric loss functions (including quadratic, linear, and logarithmic quadratic formulations) on various deep learning models such as Bidirectional LSTM (Bi-LSTM), Deep Neural Networks (DNN), and one-dimensional Convolutional Neural Network (CNN). Such asymmetric loss functions improved the prediction of the remaining useful life of an engine \cite{article24}.

%Authors in \cite{article8} proposed a model based on LSTM and RNN to predict the short-term power consumption using different hidden layers. Then, they compared between the LSTM and other methods such as SVM and Random Forest (RF) and showed that LSTM having two layers has the lowest RMSE of 11.4299. 
To predict the short-term power consumption, authors in \cite{article8} compared between the LSTM and other methods such as SVM and Random Forest (RF) and showed that LSTM with two layers achieved the lowest RMSE of 11.4299. Despite that data preprocessing was performed, they mentioned that the RMSE is still considered large due to anomalies in the data \cite{article8}. To find these anomalies in power consumption datasets, a detection method based on micro-moments and DNN was proposed in \cite{article14}, while another approach based on a mixture of polynomial regression and Gaussian distribution was introduced by \cite{article15}. In \cite{app11062742}, the DBSCAN algorithm was used in the data preprocessing stage to differentiate between the usual power consumption behaviors and the anomalous points. Then, a hybrid model that consists of a one-dimensional CNN and a Bi-LSTM was trained to get more robust predictions. This approach was tested on a dataset of households and commercial buildings and results demonstrated that it achieved a good performance in several evaluation metrics \cite{app11062742}.

So far, several attempts were considered to achieve more accurate load predictions and eliminate any anomalies. Others implemented asymmetric loss functions to reach lower underestimation errors. Yet, to the best of our knowledge, implementing both anomaly detection and asymmetric loss functions was not performed in the domain of power consumption estimation. In this work, the DBSCAN algorithm is applied on the power consumption datasets \cite{article16} to get rid of any anomalies before feeding them to several LSTM models where various asymmetric objective functions are investigated to effectively reduce underpredictions. Additionally, supplementary features are added to improve the models' load forecasting and datasets are split based on the seasonality factor.

\section{Methodology}
\label{Methodology}
In this study, we first eliminate the noisy data in the power consumption datasets after being combined with weather and calendar datasets and split into three multi-variant seasonal datasets. Then, power consumption prediction is performed using multiple LSTM algorithms with different symmetric and asymmetric loss functions in an attempt to obtain better predictions with slight underestimates in the load forecasting task. This section introduces the datasets used in this study, details the anomaly detection and substitution procedure, and presents the different implemented asymmetric loss functions.

\subsection{Datasets}
%The considered datasets are gathered from 32 European countries and provided by ENTSO-E Transparency, a European electricity data platform.
%In this paper, we chose three datasets belonging to three areas in Germany, France, and Hungary respectively. 
For this work, we chose three separate power consumption datasets belonging to three residential areas in Germany, France, and Hungary. These datasets\cite{article16} are provided by ENTSO-E Transparency, a European electricity data platform. Each dataset contains a time-series of the hourly power consumption over a five-year period (from 2015 till 2019). To consider the social and weather effects on the electric load, we added to the hourly power consumption of every area: %its corresponding weather features (temperature, direct horizontal radiation, and diffuse horizontal radiation) and two calendar features showing weekends and holidays in a Boolean representation. The weekends and holidays were labeled as 1, whereas regular weekdays were labeled as 0. 
\begin{itemize}
  \item A weather dataset\cite{article16} of its corresponding country (collected by Renewables.ninja from the NASA MERRA-2 reanalysis) with three weather features: temperature, direct horizontal radiation, and diffuse horizontal radiation.
  \item A calendar dataset of its corresponding country with two calendar features: weekends and holidays (weekends and holidays labeled as 1 and regular weekdays labeled as 0).
\end{itemize}
Additionally, to inspect the seasonal trends of the data, each dataset was divided into three seasonal datasets as follows:%, obtaining a total of six seasonal datasets. Season one starts from January until the middle of April, then season two starts from the middle of April reaching the middle of october, and finally season three represents the middle of october, November, and December. 
\begin{itemize}
  \item Season 1: January until mid-April.
  \item Season 2: mid-April until mid-October.
  \item Season 3: mid-October until the end of December.
\end{itemize}

\subsection{Anomaly Detection and Substitution}
After performing the standard preprocessing steps on each multi-variant seasonal dataset, such as normalizing the data by using the Robust Scaling method and handling the missing values, an anomaly detector based on the DBSCAN technique \cite{10.5555/3001460.3001507} was implemented to conquer the power consumption anomalies in each seasonal dataset. Given that few outliers were originally detected in the power consumption feature of the considered datasets, we manually injected additional outliers to all the seasonal datasets of Germany and Hungary to be able to prove the effectiveness of the anomaly detection approach. We considered two different cases in which the number of outliers is either equal to 1\% or 2\% of the number of data points in every seasonal dataset. Since outliers result from special events such as severe weather conditions or wrong measurements, these additional outliers were either injected on bad weather terms or on arbitrary dates (representing wrong measurements). After adding the outliers, the anomaly detection and substitution process was performed to detect these outliers in the modified seasonal datasets and replace them with the appropriate values. 

One of the advantages of the clustering algorithm DBSCAN, that groups the data points into clusters, is that it does not require to initialize the number of clusters. It only has two parameters: the maximum radius of the neighborhood eps. and the minimum number of samples in the specified eps. radius. After applying this clustering technique, if a data point did not belong to the main cluster (labeled as 0) and did not occur on a holiday, it was detected as an anomaly and was replaced by the value of the power consumption at the same day and time in the previous week. However, if the data point of the previous week was also an anomaly or if it happened to be on a holiday, we continuously shifted backwards in the same manner until reaching a data point that met the required criteria.

\subsection{Asymmetric Loss Functions}
In this work, we aim to obtain better predictions and most importantly reduce underestimations. Hence, we propose the concept of asymmetry to highly penalize underpredictions and thus reduce the underestimation error. Two asymmetric loss functions are proposed with different penalties for overestimates and underestimates.

The first asymmetric loss function $AL_1$ is designed using two different Huber-loss functions for overpredictions and underpredictions respectively. The aim of such a design is not only to highly penalize underestimates compared to overestimates, but also to more penalize the predictions with higher error values compared to the predictions with lower error values. The formulation of $AL_1$ is demonstrated as follows:
\begin{align}
{Underest\_Loss}_1&=
\begin{cases}
a\cdot|E|,\:\:   \text{if}\:\:   -1 < E \leq 0\\
a\cdot|E|^2,\:\:   \text{if}\:\:   E \leq -1
\end{cases}
,
\\
{Overest\_Loss}_1&=
\begin{cases}
b\cdot|E|^2,\:\:   \text{if}\:\:   0 < E < 1\\
b\cdot|E|,\:\:   \text{if}\:\:   E \geq 1
\end{cases}
,
\end{align}

\begin{equation}
Loss_1= \frac{\sum {Underest\_Loss}_1 + \sum {Overest\_Loss}_1}{Total \:number\: of\: estimates},
\end{equation}
where $a$ and $b$ are constants such that $a>b$ and $E$ is the difference between the predicted and the actual power consumption at a given time $t$.

The second asymmetric loss function $AL_2$ uses a linear loss function for underpredictions and an $\epsilon$-insensitive Huber-loss function for overpredictions. It also applies different penalties on overestimates and underestimates in which the penalty of underestimation is higher than the penalty of overestimation. However, all the underestimations are uniformly penalized. The twist in $AL_2$ is that it allows the overprediction errors that are less than $\epsilon_1$ to remain unchanged while penalizing higher overprediction errors non-uniformly through a Huber-loss scheme. The formulation of $AL_2$ is illustrated by:
\begin{align}
{Underest\_Loss}_2&= a\cdot|E|,\:\:   \text{if}\:\: E < 0,\\
{Overest\_Loss}_2&=
\begin{cases}
0,\:\:   \text{if}\:\:   0 \leq E < \epsilon_1\\
b\cdot|E|^2,\:\:   \text{if}\:\:   \epsilon_1 \leq E < \epsilon_2\\
b\cdot|E|,\:\:   \text{if}\:\:   E \geq \epsilon_2\
\end{cases}
,
\end{align}

\begin{equation}
Loss_2= \frac{\sum {Underest\_Loss}_2 + \sum {Overest\_Loss}_2}{Total \:number\: of\: estimates},
\end{equation}
where $a$, $b$, and $E$ are defined as previously and $\epsilon_1<\epsilon_2$.

\section{Experiments and results}
\label{results}
To assess the effectiveness of the proposed approach, we conducted several experiments on the three datasets of Germany, France, and Hungary. Every LSTM algorithm was tested on each testing set and the model’s performance was evaluated by computing the underestimation RMSE and overestimation RMSE. This section provides the experimental setup details and highlights the important results of this study. 

\subsection{Experimental Setup}
The parameters of the DBSCAN algorithm used to detect the anomalies were chosen as follows: The maximum radius of the neighborhood eps. was set to 0.11 and the minimum number of samples in the specified eps. radius was set to 3. After performing the anomaly detection and substitution procedure, every seasonal dataset was grouped into sequences using a four-width sliding window. The sequences of each dataset were then split into training and testing sets where the data from 2015 till 2018 were taken for training and the rest were taken for testing (year 2019). Then, the training sequences were fed into different LSTM algorithms to perform short-term load forecasting by outputting the hour-ahead power consumption. All the LSTM models consisted of an input layer, two LSTM and Dropout layers, and finally the dense and activation layers implemented for the output. Every model was trained for 100 epochs using Adam optimizer. This procedure was applied for all the datasets on all the LSTM models with different symmetric and asymmetric loss functions. After several hyper-parameter tuning, the parameters of $AL_1$ and $AL_2$ were set to $a=5$ and $b=2$ to insert higher weight on underestimations than overestimations. $\epsilon_1$ and $\epsilon_2$ were also heuristically chosen to be 0.005 and  0.01 respectively in $AL_2$.
At last, to inspect the effect of the asymmetric loss functions of reducing underpredictions, we compare the obtained results using an asymmetric loss to those obtained using a symmetric MSE loss function. Moreover, to assess the advantages of data splitting into three seasons, we compare all the LSTM models' performance with and without seasonality splitting. Finally, to evaluate the effectiveness of the proposed anomaly detection approach, we distinguish between the performance of the LSTM algorithms with and without applying the anomaly detection and substitution procedure. 

\subsection{Results and Discussion}
\begin{figure}[t]
    \centering
\includegraphics[scale=0.44]{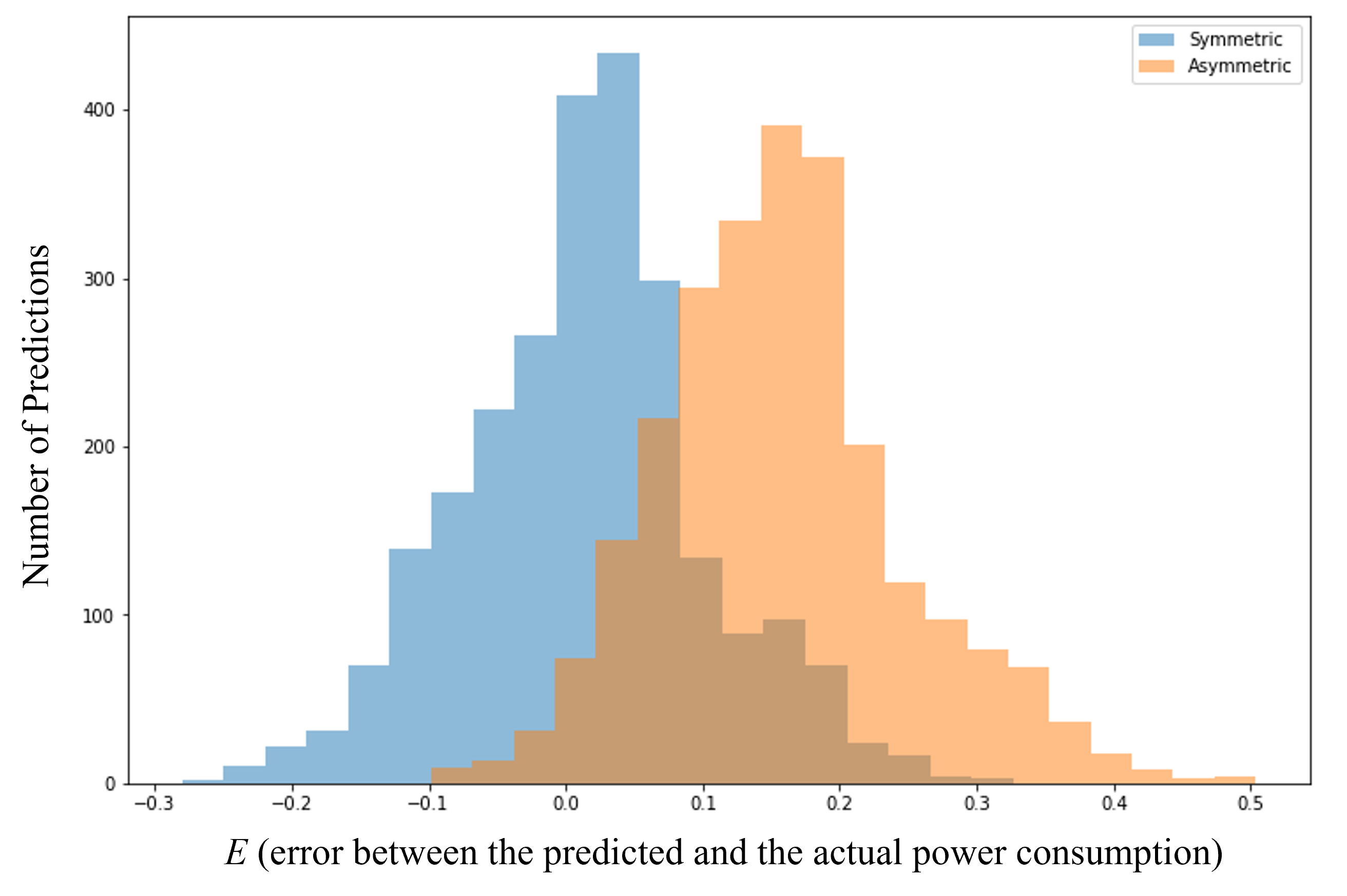}
    \caption{Histogram of the error for the symmetric loss and asymmetric loss $AL_1$ in the case of the French residence dataset in season 1.}
    \label{plot1}
\end{figure}

\begin{figure}[t]
    \centering
\includegraphics[scale=0.44]{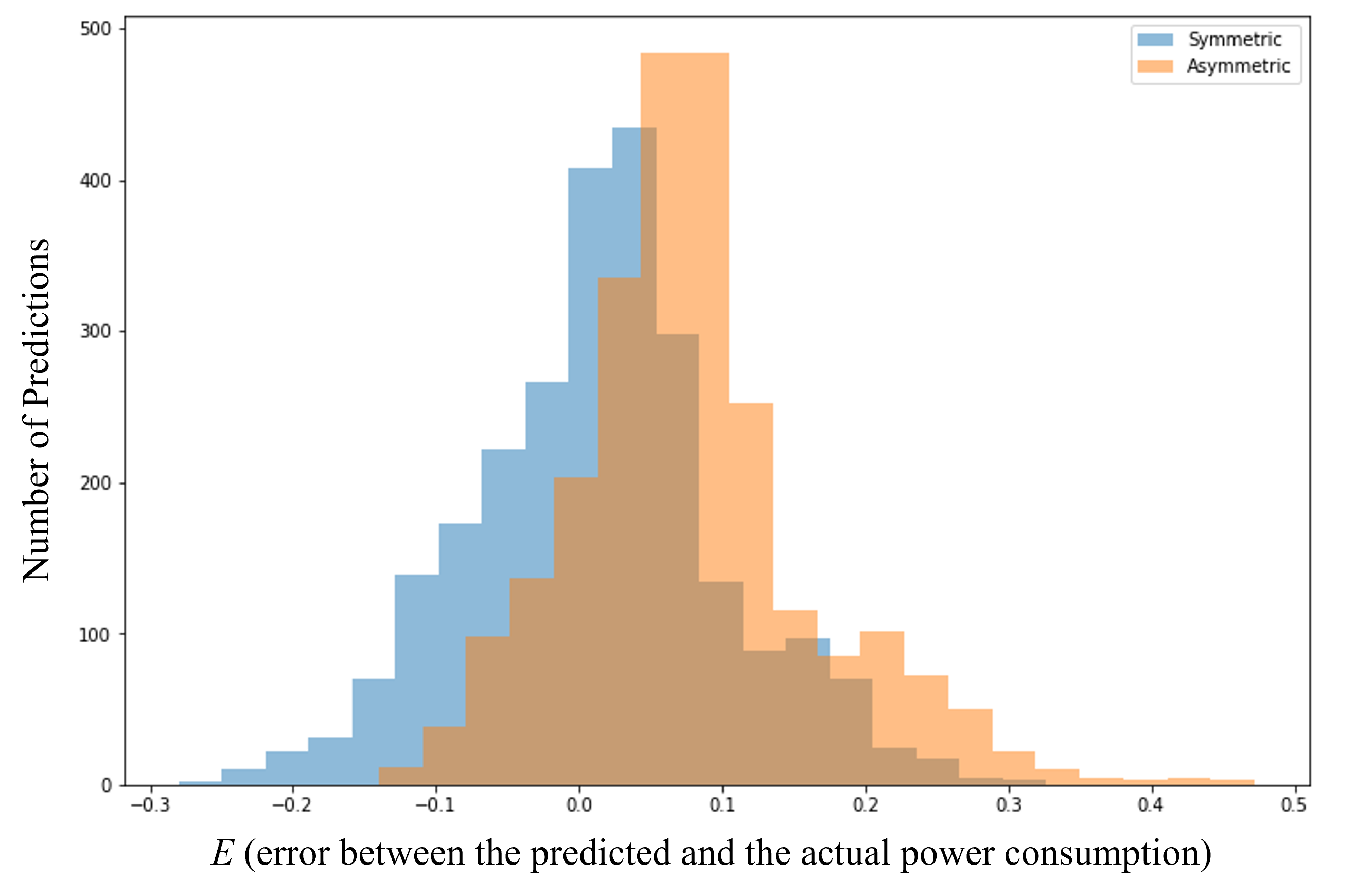}
    \caption{Histogram of the error for the symmetric loss and asymmetric loss $AL_2$ in the case of the French residence dataset in season 1.}
    \label{plot2}
\end{figure}

To examine the impact of the proposed asymmetric loss functions, the performance of each asymmetric loss function is compared to the performance of a symmetric MSE loss function. %Table \ref{table:12} averages the underestimation and overestimation errors over the three seasons when seasonality splitting is performed in the French dataset for all the different models with distinct loss functions. 
In the French dataset, the underestimation RMSE of the asymmetric loss function $AL_1$ approximately dropped by 86\%, 80\%, and 91\% in comparison to the underestimation RMSE of the symmetric loss function in seasons 1, 2, and 3 respectively. On the other hand, when using the asymmetric loss function $AL_2$, the underestimation RMSE decreased by around 63\%, 47\%, and 70\% in comparison to the underestimation RMSE of the symmetric loss in the same three seasons. Figures~\ref{plot1} and~\ref{plot2} reflect a decrease in the underestimation RMSE when using $AL_2$ and a greater decrease upon applying $AL_1$ in comparison to the symmetric loss function in the French residence in season 1. However, it can be noticed that the overestimation errors of the asymmetric loss functions are higher than the overestimation error of the symmetric loss, with $AL_1$ recording the highest overestimation error.

%\begin{table}[b]
%\begin{center}
%\begin{tabular}{ |c|c|c| } 
%\hline
% \multirow{2}{4em}{Loss function} & Underestimation &
% Overestimation \\
% & RMSE (average) & RMSE (average) \\
%\hline
% Symmetric & 0.053 & 0.054 \\ 
% $AL_1$ & 0.007 & 0.152 \\
% $AL_2$ & 0.02 & 0.094\\
%\hline
%\end{tabular}
%\end{center}
%        \caption{Underestimation and overestimation RMSE of the three loss functions averaged over the 3 seasonal datasets of the French residence.}
%    \label{table:1}
%\end{table}

Besides evaluating the effectiveness of several loss functions, the performance of the LSTM models before applying the anomaly detection technique on the seasonal datasets of Germany and Hungary was also compared to their performance after removing the anomalies in both cases of additional outliers (1\% and 2\% additional outliers). Tables \ref{table:3} and \ref{table:4} summarize the underestimation and overestimation errors of the proposed loss functions averaged over the three seasonal datasets of Germany and Hungary respectively before and after removing the anomalies in the two cases. After removing the detected anomalies in the 1\% case in the German datasets, the underestimation RMSE decreased by 56\%, 85\%, and 69\% when using the symmetric loss, $AL_1$, and $AL_2$ respectively. Similarly, the underestimation RMSE dropped by 66\%, 78\%, and 75\% when using the symmetric loss, $AL_1$, and $AL_2$ respectively in the Hungarian datasets after removing the 1\% additional anomalies. Additionally, the overestimation RMSE diminished by 47\%, 16\%, and 38\% in the German datasets and by 59\%, 64\%, and 54\% in the Hungarian datasets for the symmetric loss, $AL_1$, and $AL_2$ respectively. On the other hand, after the removal of anomalies in the 2\% case in the German datasets, the underestimation RMSE dropped by 60\%, 86\%, and 68\% and the overestimation RMSE decreased by 46\%, 21\%, and 33\% for the symmetric loss, $AL_1$, and $AL_2$ respectively. Similarly, the underestimation RMSE decreased by 72\%, 83\%, and 78\% and the overestimation RMSE dropped by 57\%, 62\%, and 57\% after removing the 2\% injected anomalies in the Hungarian datasets for the symmetric loss, $AL_1$, and $AL_2$ respectively.

\begin{table}[t]
\begin{center}
\begin{tabular}{ |c|c|c|c|c| } 
\hline
 \multirow{4}{5em}{Percentage of outliers} & \multicolumn{2}{|c|}{\text{Underestimation}} &  \multicolumn{2}{|c|}{\text{Overestimation}} \\
 & \multicolumn{2}{|c|}{\text{ RMSE (average)}} & \multicolumn{2}{|c|}{\text{ RMSE (average)}}\\
 \cline{2-5}
 & With & Without & With & Without \\
 & outliers & outliers & outliers & outliers \\
 \hline
\multicolumn{5}{|c|}{\text{Symmetric loss}}
\\ \hline
1\%   &	0.154 & 0.068 & 0.121 & \textbf{0.064} \\ 
2\%  &	0.163 & 0.066 &	0.116 &	\textbf{0.063} \\ 
\hline
\multicolumn{5}{|c|}{\text{Asymmetric loss $AL_1$}}
\\ \hline
1\% &	0.424 &	\textbf{0.062} &	0.167 &	0.141 \\
2\%  &	0.445 &	\textbf{0.063} &	0.174 &	0.138 \\
\hline
\multicolumn{5}{|c|}{\text{Asymmetric loss $AL_2$}}
\\ \hline
1\%   &	0.209 &	0.065 &	0.127 &	0.079 \\ 
2\%   &	0.206 &	0.066 &	0.128 &	0.086 \\ 
\hline
\end{tabular}
\end{center}
        \caption{Underestimation and overestimation RMSE of the three loss functions averaged over the three seasonal datasets of Germany before and after anomaly detection for the two cases of additional outliers.}
    \label{table:3}
\end{table}

\begin{table}[t]
\begin{center}
\begin{tabular}{ |c|c|c|c|c| } 
\hline
 \multirow{4}{5em}{Percentage of outliers} & \multicolumn{2}{|c|}{\text{Underestimation}} &  \multicolumn{2}{|c|}{\text{Overestimation}} \\
 & \multicolumn{2}{|c|}{\text{ RMSE (average)}} & \multicolumn{2}{|c|}{\text{ RMSE (average)}}\\
 \cline{2-5}
 & With & Without & With & Without \\
 & outliers & outliers & outliers & outliers \\
 \hline
\multicolumn{5}{|c|}{\text{Symmetric loss}}
\\ \hline
1\%   &	0.311 & 0.107 & 0.205 & \textbf{0.085} \\ 
2\%  &	0.383 & 0.106 & 0.219 & \textbf{0.095} \\ 
\hline
\multicolumn{5}{|c|}{\text{Asymmetric loss $AL_1$}}
\\ \hline
1\% &	0.584 &	\textbf{0.128} &0.373 &0.136 \\
2\%  &	0.78 &	\textbf{0.141} &0.381 &0.143 \\
\hline
\multicolumn{5}{|c|}{\text{Asymmetric loss $AL_2$}}
\\ \hline
1\%   &	0.378 &	0.093 &0.214 &0.099 \\ 
2\%   &	0.445 &	0.098 &0.253 &0.108 \\ 
\hline
\end{tabular}
\end{center}
        \caption{Underestimation and overestimation RMSE of the three loss functions averaged over the three seasonal datasets of Hungary before and after anomaly detection for the two cases of additional outliers.}
    \label{table:4}
\end{table}

We further aimed to inspect the effect of considering the seasonality factor by comparing the models' performance before and after seasonality splitting. Considering the French dataset, table \ref{table:12} shows the averaged underestimation and overestimation errors over the three seasons when seasonality splitting is performed and the underestimation and overestimation errors without seasonality splitting for the different loss functions. The reason behind choosing the French dataset for this evaluation is that the power consumption trend in France appears to be highly affected by the seasonality factor. Without seasonality splitting, the underestimation errors are 0.042, 0.009, and 0.019 when using the symmetric loss, $AL_1$, and $AL_2$ respectively. This indicates an approximate decrease of 79\% and 55\% of the underestimates when using $AL_1$ and $AL_2$ respectively in comparison to the symmetric loss. On the other hand, when performing the seasonal splitting, the underestimation RMSE approximately decreased by 87\% and 62\% on average over the three seasons when using $AL_1$ and $AL_2$ respectively in comparison to the symmetric loss. Thus, the model better learned to reduce the underestimations when splitting the data into seasons. However, the percentage increase in the overestimation error is higher after seasonality splitting in both cases of $AL_1$ and $AL_2$.

%\begin{table}[b]
%\begin{center}
%\begin{tabular}{ |c|c|c| } 
%\hline
% \multirow{2}{4em}{Loss function} & Underestimation &
% Overestimation \\
% & RMSE & RMSE \\
%\hline
% Symmetric & 0.042 & 0.052 \\ 
% $AL_1$ & 0.009 & 0.129 \\
% $AL_2$ & 0.019 & 0.087\\
%\hline
%\end{tabular}
%\end{center}
%        \caption{Underestimation and overestimation RMSE of the three loss functions without seasonality splitting of the French residence.}
%    \label{table:2}
%\end{table}

\begin{table}[t]
\begin{center}
\begin{tabular}{ |c|c|c|c|c| } 
\hline
 \multirow{4}{5em}{Loss function} & \multicolumn{2}{|c|}{\text{Underestimation}} &  \multicolumn{2}{|c|}{\text{Overestimation}} \\
 & \multicolumn{2}{|c|}{\text{ RMSE}} & \multicolumn{2}{|c|}{\text{ RMSE}}\\
 \cline{2-5}
 & With & Without & With & Without \\
 & splitting & splitting & splitting & splitting \\
\hline
Symmetric   &	0.053 & 0.042 & \textbf{0.054} & \textbf{0.052} \\ 
$AL_1$  &	\textbf{0.007} & \textbf{0.009} &	0.152 &	0.129 \\ 
$AL_2$ & 0.02 & 0.019 &	0.094 &	0.087 \\ 
\hline
\end{tabular}
\end{center}
        \caption{Underestimation and overestimation RMSE of the three loss functions with and without seasonality splitting of the French dataset.}
    \label{table:12}
\end{table}

In summary, both asymmetric loss functions succeeded in reducing the underpredictions, while noting that $AL_1$ behaved better than $AL_2$ in this manner in all the considered datasets. However, the overestimation error in the case of $AL_1$ is greater than the overestimation error in the case of $AL_2$. The reason behind such behavior could be their distinct formations: $AL_1$ greatly penalized underestimations compared to overestimations, whereas $AL_2$ less aggressively penalized underestimations. Additionally, the anomaly detection and substitution method proved to successfully reduce both underestimations and overestimations for the three loss functions, with $AL_1$ also recording the highest decrease in the underestimation error. Finally, the seasonality splitting helped in reducing underestimates while slightly increasing overestimates.

\section{Conclusion}
\label{Conclusion}
In this work, we propose two asymmetric loss functions, $AL_1$ and $AL_2$, to highly minimize the underpredictions in the power consumption estimation task using a LSTM model. The results proved that the proposed asymmetric loss functions reduced the underprediction error, with $AL_1$ achieving the best decrease. However, the overprediction error increased in both cases of $AL_1$ and $AL_2$, with $AL_1$ recording the highest increase. A clustering approach was also integrated before performing the load estimation to detect the anomalies in the power consumption data. Substituting the detected anomalies with more realistic values resulted in a better load forecast in all the LSTM models. In addition, the seasonality factor proved to help in reducing the underprediction error in the French dataset. However, the overstimation error slightly increased in both asymmetric loss functions after the seasonality splitting.

The proposed asymmetric loss functions succeeded to remarkably reduce underestimations, but failed to limit overestimations. Future work might aim to improve the formulation of these asymmetric loss functions to achieve fewer overestimations besides effectively minimizing underestimations.

\section{Acknowledgement}
\label{Acknowledgement}
This work was supported by the Maroun Semaan Faculty of Engineering and Architecture and the University Research Board at the American University of Beirut, Lebanon.
\bibliographystyle{IEEEtran}
\bibliography{sampleBibFile}

\begin{thebibliography}{10}
\providecommand{\url}[1]{#1}
\csname url@samestyle\endcsname
\providecommand{\newblock}{\relax}
\providecommand{\bibinfo}[2]{#2}
\providecommand{\BIBentrySTDinterwordspacing}{\spaceskip=0pt\relax}
\providecommand{\BIBentryALTinterwordstretchfactor}{4}
\providecommand{\BIBentryALTinterwordspacing}{\spaceskip=\fontdimen2\font plus
\BIBentryALTinterwordstretchfactor\fontdimen3\font minus
  \fontdimen4\font\relax}
\providecommand{\BIBforeignlanguage}[2]{{%
\expandafter\ifx\csname l@#1\endcsname\relax
\typeout{** WARNING: IEEEtran.bst: No hyphenation pattern has been}%
\typeout{** loaded for the language `#1'. Using the pattern for}%
\typeout{** the default language instead.}%
\else
\language=\csname l@#1\endcsname
\fi
#2}}
\providecommand{\BIBdecl}{\relax}
\BIBdecl

\bibitem{article1}
M.~Maiti, M.~Islam, and J.~Sanyal, ``Regression-based predictive models for
  estimation of electricity consumption,'' \emph{International Journal of
  Innovations in Engineering Research and Technology}, pp. 1--4, March 2021.

\bibitem{article2}
J.~T. Lalis and E.~Maravillas, ``Dynamic forecasting of electric load
  consumption using adaptive multilayer perceptron(amlp),'' in \emph{2014
  International Conference on Humanoid, Nanotechnology, Information Technology,
  Communication and Control, Environment and Management (HNICEM)}, 2014, pp.
  1--7.

\bibitem{article3}
R.~F. Berriel, A.~T. Lopes, A.~Rodrigues, F.~M. Varejão, and
  T.~Oliveira-Santos, ``Monthly energy consumption forecast: A deep learning
  approach,'' in \emph{2017 International Joint Conference on Neural Networks
  (IJCNN)}, 2017, pp. 4283--4290.

\bibitem{article8}
E.~Yuniarti, N.~Nurmaini, B.~Y. Suprapto, and M.~Naufal~Rachmatullah, ``Short
  term electrical energy consumption forecasting using rnn-lstm,'' in
  \emph{2019 International Conference on Electrical Engineering and Computer
  Science (ICECOS)}, 2019, pp. 287--292.

\bibitem{article6}
\BIBentryALTinterwordspacing
H.~Son and C.~Kim, ``A deep learning approach to forecasting monthly demand for
  residential–sector electricity,'' \emph{Sustainability}, vol.~12, no.~8,
  2020. [Online]. Available: \url{https://www.mdpi.com/2071-1050/12/8/3103}
\BIBentrySTDinterwordspacing

\bibitem{article7}
D.~L. Marino, K.~Amarasinghe, and M.~Manic, ``Building energy load forecasting
  using deep neural networks,'' in \emph{IECON 2016 - 42nd Annual Conference of
  the IEEE Industrial Electronics Society}, 2016, pp. 7046--7051.

\bibitem{article11}
\BIBentryALTinterwordspacing
M.~Torabi, S.~Hashemi, M.~R. Saybani, S.~Shamshirband, and A.~Mosavi, ``A
  hybrid clustering and classification technique for forecasting short-term
  energy consumption,'' \emph{Environmental Progress \& Sustainable Energy},
  vol.~38, no.~1, pp. 66--76, 2019. [Online]. Available:
  \url{https://aiche.onlinelibrary.wiley.com/doi/abs/10.1002/ep.12934}
\BIBentrySTDinterwordspacing

\bibitem{article22}
L.~Ehrig, D.~Atzberger, B.~Hagedorn, J.~Klimke, and J.~Döllner, ``Customizable
  asymmetric loss functions for machine learning-based predictive
  maintenance,'' in \emph{2020 8th International Conference on Condition
  Monitoring and Diagnosis (CMD)}, 2020, pp. 250--253.

\bibitem{article23}
\BIBentryALTinterwordspacing
J.~Wu, Y.~gan Wang, Y.-C. Tian, K.~Burrage, and T.~Cao, ``Support vector
  regression with asymmetric loss for optimal electric load forecasting,''
  \emph{Energy}, vol. 223, p. Article number: 119969, May 2021. [Online].
  Available: \url{https://eprints.qut.edu.au/208195/}
\BIBentrySTDinterwordspacing

\bibitem{article24}
D.~Rengasamy, B.~Rothwell, and G.~P. Figueredo, ``Asymmetric loss functions for
  deep learning early predictions of remaining useful life in aerospace gas
  turbine engines,'' in \emph{2020 International Joint Conference on Neural
  Networks (IJCNN)}, 2020, pp. 1--7.

\bibitem{article25}
J.~Zhang, Y.~Wang, and G.~Hug, ``Cost-oriented load forecasting,'' 2021.

\bibitem{article26}
M.~Stockman, R.~S. El~Ramli, M.~Awad, and R.~Jabr, ``An asymmetrical and
  quadratic support vector regression loss function for beirut short term load
  forecast,'' in \emph{2012 IEEE International Conference on Systems, Man, and
  Cybernetics (SMC)}, 2012, pp. 651--656.

\bibitem{6403624}
M.~Stockman, M.~Awad, and R.~Khanna, ``Asymmetrical and lower bounded support
  vector regression for power estimation,'' in \emph{2011 International
  Conference on Energy Aware Computing}, 2011, pp. 1--6.

\bibitem{article12}
A.~Bansal, S.~K. Rompikuntla, J.~Gopinadhan, A.~Kaur, and Z.~A. Kazi, ``Energy
  consumption forecasting for smart meters,'' 2015.

\bibitem{10.5555/3001460.3001507}
M.~Ester, H.-P. Kriegel, J.~Sander, and X.~Xu, ``A density-based algorithm for
  discovering clusters in large spatial databases with noise,'' in
  \emph{Proceedings of the Second International Conference on Knowledge
  Discovery and Data Mining}, ser. KDD'96.\hskip 1em plus 0.5em minus
  0.4em\relax AAAI Press, 1996, p. 226–231.

\bibitem{article16}
\BIBentryALTinterwordspacing
``A platform for open data of the european power system.'' [Online]. Available:
  \url{https://open-power-system-data.org/}
\BIBentrySTDinterwordspacing

\bibitem{article14}
Y.~Himeur, A.~Alsalemi, F.~Bensaali, and A.~Amira, ``A novel approach for
  detecting anomalous energy consumption based on micro-moments and deep neural
  networks,'' \emph{Cognitive Computation}, vol.~12, 11 2020.

\bibitem{article15}
\BIBentryALTinterwordspacing
W.~Cui and H.~Wang, ``A new anomaly detection system for school electricity
  consumption data,'' \emph{Information}, vol.~8, no.~4, 2017. [Online].
  Available: \url{https://www.mdpi.com/2078-2489/8/4/151}
\BIBentrySTDinterwordspacing

\bibitem{app11062742}
\BIBentryALTinterwordspacing
F.~Ünal, A.~Almalaq, and S.~Ekici, ``A novel load forecasting approach based
  on smart meter data using advance preprocessing and hybrid deep learning,''
  \emph{Applied Sciences}, vol.~11, no.~6, 2021. [Online]. Available:
  \url{https://www.mdpi.com/2076-3417/11/6/2742}
\BIBentrySTDinterwordspacing

\end{thebibliography}

\vspace{12pt}

\end{document}